\documentclass[twoside,11pt]{article}

\usepackage[accepted]{melba2} 
\usepackage{amsmath,amsfonts}

\usepackage{todonotes}
\usepackage{booktabs}
\usepackage{verbatim}
\usepackage{multirow}
\usepackage{amsmath,upgreek,bm}
\usepackage{enumitem}
\usepackage{adjustbox}
\usepackage{amssymb}
\usepackage{pifont}
\newcommand{\tr}[1]{{#1}^\top}

\definecolor{myrevisioncolor}{rgb}{0, 0, 0}
\definecolor{mydiscussioncolor}{rgb}{0, 0, 0}

\newcommand{\revision}[1]{\textcolor{myrevisioncolor}{#1}}

\newcommand{\mr}[1]{\mathrm{#1}}

\DeclareMathOperator{\E}{\mathbb{E}}

\newcommand{\MI}{I} 
\newcommand{\Data}{\mathcal{D}}
\newcommand{\TransSet}{\mathcal{T}}

\newcommand{\img}{\mathbf{x}}
\newcommand{\lbl}{\mathbf{y}}
\newcommand{\Int}{\mathbb{Z}}
\newcommand{\Real}{\mathbb{R}}
\newcommand{\params}{\bm{\uptheta}}
\newcommand{\loss}{\mathcal{L}}
\newcommand{\lossSup}{\loss_{\mathrm{spv}}}
\newcommand{\lossMI}{\loss_{\mathrm{MI}}}

\newcommand{\lossCons}{\loss_{\mathrm{cons}}}

\newcommand{\neigh}{\mathcal{N}}

\newcommand{\cmark}{\ding{51}}%

\newcommand{\parenc}{\params_{\mr{enc}}}

\newcommand{\PP}{P}

\newcommand{\enc}{\phi_{\mr{enc}}} 
\newcommand{\dec}{\phi_{\mr{dec}}}

\newcommand{\paperurl}{\url{https://github.com/jizongFox/MI-based-Regularized-Semi-supervised-Segmentation}}

\newcommand{\acdc}[1]{
\includegraphics[ width=0.163\linewidth]{figures/visual_inspection_new/acdc/case#1/gt.png} &
\includegraphics[ width=0.163\linewidth]{figures/visual_inspection_new/acdc/case#1/ps.png} &
\includegraphics[ width=0.163\linewidth]{figures/visual_inspection_new/acdc/case#1/iic.png} &
\includegraphics[ width=0.163\linewidth]{figures/visual_inspection_new/acdc/case#1/uda.png} &
\includegraphics[ width=0.163\linewidth]{figures/visual_inspection_new/acdc/case#1/mt.png} &
\includegraphics[ width=0.163\linewidth]{figures/visual_inspection_new/acdc/case#1/ours.png} 
}

\newcommand{\prostate}[1]{
\includegraphics[ width=0.163\linewidth]{figures/visual_inspection_new/prostate/case#1/gt.png} &
\includegraphics[ width=0.163\linewidth]{figures/visual_inspection_new/prostate/case#1/ps.png} &
\includegraphics[ width=0.163\linewidth]{figures/visual_inspection_new/prostate/case#1/iic.png} &
\includegraphics[ width=0.163\linewidth]{figures/visual_inspection_new/prostate/case#1/uda.png} &
\includegraphics[ width=0.163\linewidth]{figures/visual_inspection_new/prostate/case#1/mt.png} &
\includegraphics[ width=0.163\linewidth]{figures/visual_inspection_new/prostate/case#1/ours.png} 
}

\newcommand{\spleen}[1]{
\includegraphics[ width=0.163\linewidth]{figures/visual_inspection_new/spleen/case#1/gt.png} &
\includegraphics[ width=0.163\linewidth]{figures/visual_inspection_new/spleen/case#1/ps.png} &
\includegraphics[ width=0.163\linewidth]{figures/visual_inspection_new/spleen/case#1/iic.png} &
\includegraphics[ width=0.163\linewidth]{figures/visual_inspection_new/spleen/case#1/uda.png} &
\includegraphics[ width=0.163\linewidth]{figures/visual_inspection_new/spleen/case#1/mt.png} &
\includegraphics[ width=0.163\linewidth]{figures/visual_inspection_new/spleen/case#1/ours.png}
}

\newcommand{\mmwhs}[1]{
\includegraphics[ width=0.163\linewidth]{figures/visual_inspection_new/mmwhs/case#1/gt.png} &
\includegraphics[ width=0.163\linewidth]{figures/visual_inspection_new/mmwhs/case#1/ps.png} &
\includegraphics[ width=0.163\linewidth]{figures/visual_inspection_new/mmwhs/case#1/iic.png} &
\includegraphics[ width=0.163\linewidth]{figures/visual_inspection_new/mmwhs/case#1/uda.png} &
\includegraphics[ width=0.163\linewidth]{figures/visual_inspection_new/mmwhs/case#1/mt.png} &
\includegraphics[ width=0.163\linewidth]{figures/visual_inspection_new/mmwhs/case#1/ours.png}
}

\newcommand{\showfeaturespace}[1]{
\includegraphics[ width=0.24\linewidth]{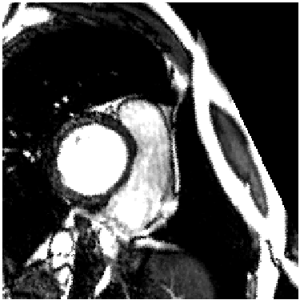} &
\includegraphics[ width=0.24\linewidth]{figures/feature_space/3_#1_feature2.png} &
\includegraphics[ width=0.24\linewidth]{figures/feature_space/3_#1_feature1.png} &
\includegraphics[ width=0.24\linewidth]{figures/feature_space/3_#1_prediction.png} 
}

\melbaheading{009}{https://www.melba-journal.org/article/23700}{2021}{1-29}{10/2020}{5/2021}{Peng, Pdersoli and Desrosiers}{Medical Imaging with Deep Learning
          (MIDL) 2020}{Marleen de Bruijne, Tal Arbel, Ismail Ben Ayed, Hervé Lombaert}
\ShortHeadings{Semi-supervised Segmentation with Mutual Information Regularization}{Peng, Pdersoli and Desrosiers}
\firstpageno{1}

\title{Boosting Semi-supervised Image Segmentation with Global and Local Mutual Information Regularization
}

\author{\name Jizong Peng \email jizong.peng.1@etsmtl.net \\ 
	\addr Department of Software and IT Engineering, ETS Montreal, Canada
	\AND
	\name Marco Pedersoli \email marco.pedersoli@etsmtl.ca \\
	\addr Department of Systems Engineering, ETS Montreal, Canada
	\AND
	\name Christian Desrosiers \email christian.desrosiers@etsmtl.ca \\
	\addr Department of Software and IT Engineering, ETS Montreal, Canada
}

\begin{document}

\maketitle

\begin{abstract}
%

\revision{The scarcity of labeled data often impedes the application of deep learning to the segmentation of medical images}. Semi-supervised learning seeks to overcome this limitation by exploiting unlabeled examples in the learning process. In this paper, we present a novel semi-supervised segmentation method that leverages mutual information (MI) on categorical distributions to achieve both global representation invariance and local smoothness. In this method, we maximize the MI for intermediate feature embeddings that are taken from both the encoder and decoder of a segmentation network. We first propose a global MI loss constraining the encoder to learn an image representation that is invariant to geometric transformations. Instead of resorting to computationally-expensive techniques for estimating the MI on continuous feature embeddings, we use projection heads to map them to a discrete cluster assignment where MI can be computed efficiently. Our method also includes a local MI loss to promote spatial consistency in the feature maps of the decoder and provide a smoother segmentation. Since mutual information does not require a strict ordering of clusters in two different assignments, we incorporate a final consistency regularization loss on the output which helps align the cluster labels throughout the network. We evaluate the method on four challenging publicly-available datasets for medical image segmentation. Experimental results show our method to outperform recently-proposed approaches for semi-supervised segmentation and provide an accuracy near to full supervision while training with very few annotated images 
\end{abstract}


\section{Introduction}

Supervised learning approaches based on deep convolutional neural networks (CNNs) have achieved outstanding performance in a wide range of segmentation tasks. However, such approaches typically require a large amount of labeled images for training. In medical imaging applications, obtaining this labeled data is often expensive since annotations must be made by trained clinicians, typically in 3D volumes, and regions to segment can have very low contrast. Semi-supervised learning is a paradigm which reduces the need for fully-annotated data by exploiting the abundance of unlabeled data, i.e. data without expert-annotated ground truth. In contrast to standard approaches that learn exclusively from labeled data, semi-supervised methods also leverage intrinsic properties of unlabeled data (or \textit{priors}) to guide the learning process. 

Among the main approaches for semi-supervised segmentation, those employing consis\-ten\-cy-based regularization and unsupervised representation learning have shown a great potential at exploiting unlabeled data \revision{\citep{perone2018deep,perone2019unsupervised, bortsova2019semi, li2018semi, chaitanya2020contrastive}}. The former approach, which leverages the principle of transformation equivariance, \revision{i.e., $f(T(x)) = T(f(x))$ for a geometrical transformation $T$}, enforces the segmentation network to predict similar outputs for different transformed versions of the same unlabeled image \citep{perone2018deep, bortsova2019semi, li2018semi}. Typical geometrical transformations include small translations, rotations or scaling operations on the image. A common limitation for consistency-based methods, however, is that they ignore the dense and structured nature of image segmentation, and impose consistency on different pixels independently. On the other hand, representation learning \citep{bengio2013representation} uses unlabeled data in a pre-training step to find an internal representation of images (i.e., convolutional feature maps) which is useful to the downstream analysis task. A recent technique based on this paradigm is contrastive learning \citep{oord2018representation, tian2019contrastive}. In this technique, a network is trained with a set of paired samples from the same joint distribution (positive pair) or different distributions (negative pair). A contrastive loss is employed to make the representation of positive-pair images similar to each other, and the representation of negative-pair images to be different. Despite showing encouraging results for segmentation 
\citep{chaitanya2020contrastive}, contrastive learning methods typically suffer from major drawbacks. In particular, they require \revision{a large} number of negative pairs and a large batch size to work properly \revision{\citep{chen2020simple}}, which makes training computationally expensive for medical image segmentation. These drawbacks are primarily due to the use of a continuous-variable representation that \revision{makes the estimation of the joint distribution of samples or their mutual information more difficult} \citep{poole2019variational, ji2018iic}. 

An alternative approach to unsupervised representation learning, based on a discrete representation, 
is clustering~\revision{\citep{ji2018iic,caron2018deep,peng2019information}}. In deep clustering, a network is trained with unlabeled data to map examples with similar semantic meaning to the same cluster label. The challenge of this unsupervised task is twofold. Firstly, using traditional pairwise similarity losses like KL divergence or $L_2$ leads to the trivial solution where all examples are mapped to the same cluster \citep{bridle1992unsupervised,krause10discr,hu2018imsat,ji2018iic}. Also, unlike for supervised classification, the labels in clustering are arbitrary and any permutation of these labels gives an equivalent solution. To address these challenges, \cite{ji2018iic} recently proposed an Information Invariant Clustering (IIC) algorithm based on mutual information (MI). The MI between two variables $X$ and $Y$ corresponds to the KL divergence between their joint distribution and the product of their marginal distributions: 
\begin{equation}\label{eq:mi-kl}
 \MI(X;Y) \ = \ D_{\mr{KL}}\big(p(X,Y)\, || \, p(X)\,p(Y)\big).
\end{equation}
Alternatively, MI can also be defined as the difference between the entropy of $Y$ and its entropy conditioned on $X$:
\begin{equation}\label{eq:mi-entropy}
\begin{aligned}
 \MI(X;Y) \ = \ & H(Y) \, - \, H(Y|X) \\
 \ = \ & \E_Y\big[\log \E_X[\,p(Y|X)\,]\,\big] \, - \, \E_{X,Y}[\,\log p(Y|X)\,].
 \end{aligned}
\end{equation}
The IIC algorithm seeks network parameters which maximize the MI between the cluster labels of different transformed versions of an image. As can be seen from Eq. (\ref{eq:mi-entropy}), \revision{if $X$ is a random variable corresponding to an image and $Y$ is another variable representing a cluster label}, this approach avoids the trivial assignment of all images the same cluster since the first term (entropy) is maximized for uniformly distributed clusters $Y$~\citep{hu2018imsat,zhao2019region}. 

Recently, \cite{peng2020mutual} adapted the IIC algorithm to semi-supervised segmentation. In their work, a network is trained with both labeled and unlabeled data such that its prediction for labeled images is similar to the ground-truth mask, and output labels for neighbor patches in different transformed versions of the same unlabeled image (after reversing the transform) have a high MI. This MI-based approach has two positive effects on segmentation. First, it makes the network more robust to image variability corresponding to the chosen transformations. Second, it increases the local smoothness of the segmentation and \revision{avoids collapse} to a single class. Since MI is invariant to the permutation of cluster labels, another loss based on KL is also added to align these labels across different image patches during training. Although leading to improved performance for the various segmentation tasks, this recent method has the following two limitations: 1) it only regularizes the output of the network, not its internal representation; 2) the regularization is only applied locally in the image, and not globally. 

\paragraph{Contributions} In this paper, we propose a novel semi-supervised segmentation method which uses the MI between representations computed at different hierarchical levels of the network to regularize its prediction both globally and locally. The proposed method employs auxiliary projection heads on layers of both the encoder and the decoder to group together feature vectors that are semantically related. Two separate strategies are used to achieve global and local regularization. In the global regularization strategy, we consider the entire feature map at a given layer as a representation of the input image and learn a mapping from this representation to a set of cluster labels. By maximizing the MI between the cluster assignments of two transformed versions of the same image, we thus promote invariance (equivariance) of the network with respect to the considered transformations. On the other hand, the local regularization strategy learns clusters for each spatial location of feature maps in the decoder, and maximizes the MI between cluster assignments of two neighbor feature vectors in transformed images. This enhances the spatial consistency of the segmentation output. 

The detailed contributions are as follows:
\begin{itemize}[itemsep=1pt,topsep=2pt]
\item We propose the first semi-supervised segmentation method using MI maximization on categorical labels to achieve both global representation invariance and local smoothness. Our method is orthogonal to state-of-the-art consistency-based approaches like Mean Teacher which impose consistency only on the output space. 
By clustering feature embeddings from different hierarchical levels and scales, our method can effectively achieve a higher performance with very few labeled images. 


\item This paper represents a major extension of our previous work in \citep{peng2020mutual} where clustering-based MI regularization was only applied locally on the network output. In contrast, the method proposed in this paper maximizes MI between both local and global feature embeddings from different layers of the network encoder and decoder. In a comprehensive set of experiments, we show that feature representations from separate hierarchical levels capture complementary information and contribute differently to performance. Moreover, we visually demonstrate the clustering effect of the proposed loss that maximizes MI between categorical labels. 
\end{itemize}

The rest of this paper is as follows. In the next section, we give a summary of related work on semi-supervised segmentation and unsupervised representation learning. In Section \ref{sec:method}, we then present the proposed semi-supervised segmentation method and explain how MI between cluster assignment labels is leveraged to achieve both local and global segmentation consistency. Our comprehensive experimental setup, involving four challenging segmentation datasets and comparing against strong baselines, is detailed in Section \ref{sec:experiments}. Results, reported in Section \ref{sec:results}, show our method to significantly outperform compared approaches and yield performance near to full supervision when trained with only 5\% of labeled examples.

\section{Related works}

\paragraph{Semi-supervised segmentation}

Although initially developed for classification~\citep{Oliver2018Realistic}, a wide range of semi-supervised methods have also been proposed for semantic segmentation. These methods are based on various learning techniques, including self-training~\citep{bai2017semi}, distillation~\citep{radosavovic2018data}, attention learning~\citep{min2018robust}, adversarial learning~\citep{souly2017semi,zhang2017deep}, entropy minimization~\citep{vu2019advent}, co-training~\citep{peng2019deep,zhou2019semi}, temporal ensembling~\citep{perone2018deep}, manifold learning~\citep{baur2017semi}, and data augmentation~\citep{chaitanya2019semi,zhao2019data}. Among recently proposed methods, consistency-based regularization has emerged as an effective way to improve performance by enforcing the network to output similar predictions for unlabeled images under different transformations~\citep{bortsova2019semi}. Following this line of research, the $\Pi$ model perturbs an input image with stochastic transformations or Gaussian noise and improves the generalization of a network by minimizing the discrepancy of its output for perturbed images. Virtual adversarial training (VAT) replaces the random perturbation with an adversarial one targeted at fooling the trained model. By doing so, the network efficiently learns a local smoothness prior and becomes more resilient to various noises. Consistency has also been a key component in temporal ensembling techniques like Mean Teacher~\citep{perone2018deep}, where the output of a student network at different training iterations is made similar to that of a teacher network whose parameters are an exponential weighted temporal average of the student's. This method has shown great success for various semi-supervised tasks such as brain lesion segmentation \citep{cui2019semi}, spinal cord gray matter segmentation \citep{perone2018deep} and left atrium segmentation \citep{yu2019uncertainty}. 

Despite improving performance in semi-supervised settings, a common limitation of the above methods is that they consider the prediction for different pixels as independent and apply a pixel-wise distance loss such as KL divergence or $L_2$ loss. This ignores the dense structure nature of the segmentation. Moreover, those approaches only regularize the output of the network for perturbed inputs, ignoring the hierarchical and multi-scale information found in different layers of the network.

\begin{sloppypar}
\paragraph{Unsupervised representation learning}
\label{sec:unsupervised_representation}

Important efforts have also been invested towards learning robust representations from unlabeled data. In self-supervised learning~\citep{noroozi2016unsupervised, kim2018learning, noroozi2018boosting}, unlabeled data are typically exploited in a first step to learn a given pretext task. This pretext task helps the network capture meaningful representations that can improve learning downstream tasks like classification or segmentation with few labeled data. \cite{taleb2019multimodal} trained a convolutional network to solve jigsaw puzzles and used the learned representation to boost performance for multi-modal medical segmentation. Other pretext jobs include predicting the transformation applied to an input image \citep{zhang2019aet,wang2019enaet} and converting a grey-scale image to RGB \citep{zhang2016colorful}. 
\end{sloppypar}

Recently, contrastive learning was shown to be an effective strategy for semi-supervised learning. In this approach, one trains a network with a set of paired examples, together with a critic function to tell whether a pair of examples comes from their joint distribution or not. \revision{In their Contrastive Predicted Coding (CPC) approach, \cite{oord2018representation} use a contrastive loss to learn a representation which can be predicted with an autoregressive model.} \cite{tian2019contrastive} proposed a Contrastive Multiview Coding (CMC) method where the network must produce similar features for images of different modalities if they correspond to the same object. \cite{chen2020simple} instead learn to predict whether a pair of images comes from a same image under different data augmentations. So far, only a single work has investigated contrastive learning for medical image segmentation~\citep{chaitanya2020contrastive}. In this work, a network is trained to distinguish whether a pair of 2D images comes from the same physical position of their corresponding 3D volumes or not. Although contrastive learning has been shown to be related to MI~\citep{tian2019contrastive}, the approach of \cite{chaitanya2020contrastive} differs significantly from our method. First, their approach uses a standard contrastive loss between continuous vectors that requires sampling a large number of negative pairs and is expensive for image segmentation. In contrast our method exploits the MI between categorical labels, which can be computed efficiently. Moreover, whereas they impose consistency between corresponding positions in two different feature maps, our method also enforces it between neighbor positions and for different image transformations. This adds local smoothness to the feature representations and helps generate a more plausible segmentation. Last, whereas their approach only leverages unlabeled data in a pre-training step, we optimize the segmentation network with both labeled and unlabeled images in a single step.

Deep clustering has also been explored to learn robust representation of image data. Since it favors balanced clusters, thus avoiding the collapse of the solution to a single cluster, and does not make any assumption about the data distribution, MI has been at the core of several deep clustering methods. \revision{One of them}, Information Maximizing Self-Augmented Training (IMSAT) \citep{hu2018imsat}, maximizes the MI between input data $X$ and the cluster assignment $Y$. The output is regularized through the use of virtual adversarial samples \citep{miyato2019virtual}, imposing that the original sample and the adversarial one should have a similar cluster assignment probability distribution. A related approach, called Invariant Information Clustering (IIC) \citep{ji2018iic}, instead maximizes the MI between cluster assignments of a sample and its transformed versions. Recently, \cite{peng2020mutual} proposed an semi-supervised segmentation method inspired by IIC which encourages nearby patches in the network's output map, for two transformed versions of the same unlabeled image, to have a high MI. As mentioned above, this avoid the trivial assignment of all pixels to a single class and also promotes spatial smoothness in the segmentation. However, a common limitation of deep clustering methods for image classification and segmentation is that they only consider the network output, and ignore the rich semantic information of features inside the network. 

\paragraph{Estimating MI}
\label{sec: MI_estimation}

\revision{Capturing MI between two random variables is a difficult task, especially when these variables are continuous and/or high-dimensional. Traditional density- or kNN-based methods \citep{suzuki2008approximating, vejmelka2007mutual} do not scale well to complex data such as raw images. Recently, variational approaches have become popular for estimating MI between latent representations and observations \citep{hjelm2018learning, oord2018representation} or between two related latent representations \citep{tian2019contrastive, chaitanya2020contrastive}. These approaches instead maximize a variational lower bound to MI, thus making the problem tractable. Related to our work, \cite{belghazi2018mine} leveraged the dual representation of KL divergence to develop a variational neural MI estimator (MINE) for image classification. In their Deep InfoMax method, \cite{hjelm2018learning} used MINE to measure and maximize the MI between global and local representations. Various improvements have later been proposed to mitigate the high estimation variance of MINE \citep{mcallester2020formal}, such as using $f$-divergence representation~\citep{nowozin2016f}, Jensen–Shannon (JS) divergence based optimization~\citep{hjelm2018learning, zhao2020deep}, and clipping output with a prefixed range~\citep{song2019understanding}. Contrastive-based methods have been shown to underestimate MI~\citep{hjelm2018learning, mcallester2020formal} and  require a large number of negative examples \citep{tian2019contrastive}. As alternative to MINE, discriminator-based MI estimation \citep{liao2020demi, mukherjee2020ccmi} trains a \textit{binary} classification network to  directly emulate the density ratio between the joint distribution and the product of marginal.}

\revision{Our method differs significantly from the above-mentioned approaches. First, these approaches usually define a \textit{statistic network} \citep{belghazi2018mine} or a discriminator~\citep{liao2020demi, mukherjee2020ccmi} to project high dimension data to a scalar, which often consists of convolution and MLP layers~\citep{hjelm2018learning, liao2020demi}. {On the contrary, our method employs a simple classifier to find proper categorical distributions and then maximize the estimated MI. This helps optimize the mutual information between dense representations efficiently}. Compared to contrastive-based methods \citep{tian2019contrastive, chaitanya2020contrastive}, as we will show in Sec. \ref{sec: impact_k}, we can improve performance by simply increasing the number of clusters $K$ instead of the batch size. The latter is not easily achieved in a memory- and computation-expensive task like segmentation. Last but not least, above-mentioned approaches rely on sampling \textit{both} positive and negative pairs and seek to identify a \textit{binary} decision boundary separating the joint distribution from the product of marginals. In contrast, we do not require negative pairs, similar to the recently proposed BYOL method~\citep{grill2020bootstrap}, but instead learn a fine-grain \textit{multi-class} mapping. We leave as future work the comparison of different MI estimation strategies for semi-supervised segmentation.}

\section{Proposed method}\label{sec:method}

\begin{figure}[t!]
{\includegraphics[width=0.99\linewidth]{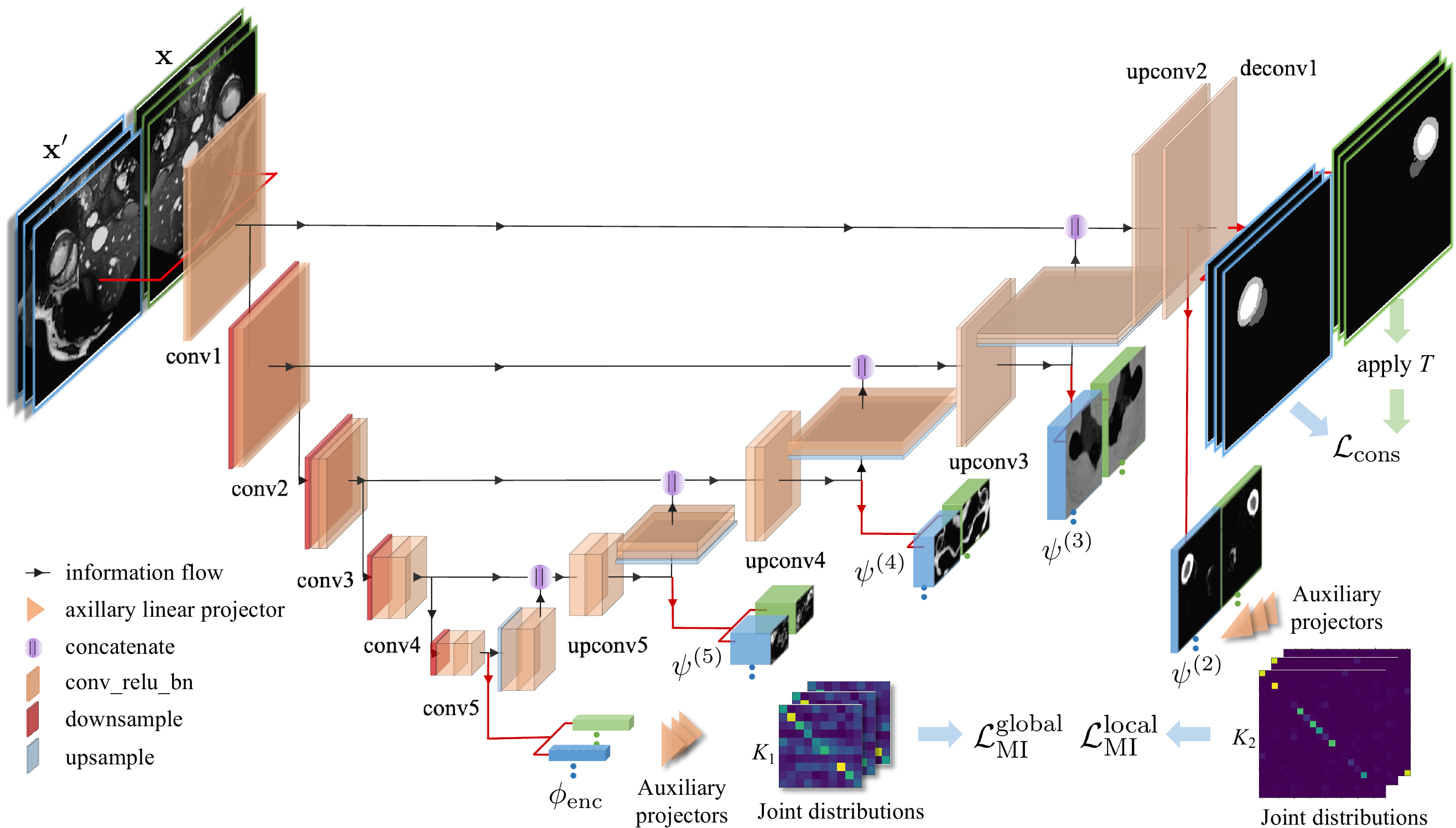}}
\caption{\textbf{Training pipeline of our semi-supervised segmentation method}. Given an unlabeled image $\img$ and its transformation $\img'$, we seek to maximize the mutual information of their intermediate feature representation with the help of auxiliary projectors. We maximize the global MI ($\lossMI^{\mr{global}}$ loss) for embeddings taken from the encoder to learn transformation-invariant representation. Meanwhile, local MI is maximized ($\lossMI^{\mr{local}}$ loss) for embeddings taken from the decoder, encouraging the network to group schematically-related regions while taking into consideration the spatial smoothness. $\lossCons$ further enforces the consistency on prediction distributions through different transformation and ensures the alignment of cluster label throughout the network. 
}
\label{fig:diagram} 
\end{figure}

We start by defining the problem of semi-supervised segmentation considered in this work and give and overview of the proposed method. We then explain each component of our method in greater details.

\subsection{Semi-supervised segmentation model} 

We consider a semi-supervised segmentation task where we have a labeled dataset $\Data_l$ of image-label pairs ($\img, \lbl$), with image $\img \in \Real^\Omega$ and ground-truth labels $\lbl \in \{1,\ldots,C\}^\Omega$, and a larger unlabeled dataset $\Data_{u}$ consisting of images without their annotations. Here, $\Omega=\{1,\ldots,W\} \times \{1,\ldots,H\}$ represents the image space (i.e., set of pixels) and $C$ is the number of segmentation classes. We seek to learn a neural network $f$ parametrized by $\params$ to predict the segmentation label of each pixel of the input image.

Fig. \ref{fig:diagram} illustrates the proposed network architecture and training pipeline. We use an encoder-decoder architecture for the segmentation network, where encoder $\enc$ extracts the information of an input image $\img$ by passing it through multiple convolutional blocks with down-sampling, and squeezes it into a compact embedding $\enc(\img)$. This embedding usually summarizes the global context of the image. The decoder $\dec$ then gradually up-samples this embedding, possibly using some side information, and outputs the prediction $\lbl = \dec(\enc(\img))$. While our method is agnostic to the choice of segmentation network, we consider in this work the well-known U-Net architecture \citep{ronneberger2015u} which achieved good performance on various bio-medical segmentation tasks. Compared to traditional encoder-decoder architectures, U-Net adds skip connections from the encoder to the decoder to reuse feature maps of same resolution in the decoder, thus helping to preserve fine details in the segmentation.

Following the main stream of semi-supervised segmentation approaches, our method exploits both labeled and unlabeled data during training. The parameters $\params$ of the network are learned by optimizing the following loss function:
\begin{equation}\label{eq:totalLoss}
\loss(\params; \Data_l, \Data_u) \ = \ \lossSup(\params; \Data_l) \ + \ \lambda_1\lossMI^{\mr{global}}(\params; \Data_u) \ + \ \lambda_2\lossMI^{\mr{local}}(\params; \Data_u) \ + \ \lambda_3\lossCons(\params; \Data_u).
\end{equation}
This loss is comprised of four separate terms, which relate to different aspects of the segmentation and whose relative importance is controlled by hyper-parameters $\lambda_1, \lambda_2, \lambda_3 \geq 0$. As in standard supervised methods, $\lossSup$ uses labeled data $\Data_l$ and imposes the pixel-wise prediction of the network for an annotated image to be similar to the ground truth labels. While other segmentation losses like the Dice loss could also be considered, our method uses the well-known cross-entropy loss:
\begin{equation}\label{eq:cross-entropy}
 \lossSup(\params; \Data_l) \ = \ - \, \frac{1}{|\Data_l| \,|\Omega|}\!\sum_{(\img,\lbl) \in \Data_l} \sum_{(i,j)\in\Omega} y_{ij} \log f_{ij}(\img; \params).
\end{equation}

Since we have no annotations for images in $\Data_u$, we instead use this unlabeled data to regularize the learning and guide the optimization process toward good solutions. This is achieved via three loss terms: $\lossMI^{\mr{global}}$, $\lossMI^{\mr{local}}$, and $\lossCons$. The first two are based on maximizing the MI between the feature embeddings of an image under different data augmentation, where embeddings can come from different hierarchical levels of both the encoder and the decoder. Specifically, we want to capture the information dependency between the semantically-related feature maps, while avoiding the complex computation of this dependency in continuous feature space. To obtain an accurate and efficient estimation of MI, we resort to a set of auxiliary projectors that convert features into categorical distributions. 

We exploit this idea in two complementary regularization losses, focusing on global MI and local MI. The global MI loss $\lossMI^{\mr{global}}$ considers the embedding $\enc(\img)$ produced by the encoder as a global representation of an image $\img$, and enforces this representation to preserve its information content under a given set of image transformations. On the other hand, the local MI loss $\lossMI^{\mr{local}}$ is based on the principle that information within a small region of the image should be locally invariant. That is, the MI between a vector in a feature map and its neighbor vectors should be high, if they correspond to the same semantic region of the image. By maximizing the MI between neighbor vectors, we can thus obtain feature representations and a segmentation output which are spatially consistent. 

The last term in (\ref{eq:totalLoss}), $\lossCons$, is a standard transformation consistency regularizer that is included for two main reasons. First, as in regular consistency-based methods, it forces the network to produce the same pixel-wise output for different transformations of a given image, after reversing the transformations. Therefore, it directly promotes equivariance in the network. The second reason stems from the fact that MI is permutation-invariant and, thus, any permutation of labels in two cluster assignments does not change their MI. Hence, $\lossCons$ helps align those labels across the network. We note several differences between $\lossCons$ and $\lossMI^{\mr{local}}$. While $\lossCons$ is only employed at the network output, $\lossMI^{\mr{local}}$ may also be used at different layers of the decoder. Moreover, because it imposes strict equality, $\lossCons$ can only be used between corresponding pixels in two images. In contrast, $\lossMI^{\mr{local}}$ also considers information similarity between feature map or output locations that are not in perfect correspondence. In the following subsections, we present each of the three regularization loss terms individually. 


\subsection{Global mutual information loss}

Let $\img$ be an image sampled from $\Data_u$ and $T$ an image transformation drawn from a transformation pool $\TransSet$. Transformation $T$ is typically a random crop, horizontal flip, small rotation, or a combination of these operations. After applying $T$ on $\img$, the transformed image $\img'=T(\img)$ should share similar contextual information as $\img$. Consequently, we expect a high MI between random variables corresponding to original and transformed images. Based on this idea, we want the encoder $\enc$ to learn latent representations for these images which maximizes their mutual information:
\begin{equation}
 \max_{\parenc} \ \MI\big(\enc(X); \, \enc(X')\big)
 \label{equ:max_encoding}
\end{equation}
where $\parenc$ are the enconder's learnable parameters. However, optimizing directly Eq. (\ref{equ:max_encoding}) is notoriously difficult as the two variables are in continuous space. For instance, one has to learn a critic function and maximize a variational lower bound of $\MI$, which may result in heavy computation and high variance~\citep{liao2020demi, song2019understanding}. 

To overcome this problem, we adapt the method proposed for unsupervised clustering and project the embeddings into categorical distributions $p(Z \, | \, \img) =  g(\enc(\img))) \in [0,1]^K$ with an auxiliary projector $g$ consisting of a linear layer followed by a softmax activation. Using this approach, embeddings $\enc(\img)$ and $\enc(\img')$ are converted to cluster probability distributions $p(Z \, | \, \img)$ and $p(Z \, | \, \img')$ with a predefined cluster number $K$. This projection introduces a bottleneck effect on (\ref{equ:max_encoding}) since
\begin{equation}
 \MI\big(g(\enc(X)); \, g(\enc(X'))\big) \, \leq \, \MI\big(\enc(X); \, \enc(X')\big)
 \label{equ:max_probs}
\end{equation}
\revision{The information bottleneck theory states that a capacity-limited network $g$ can lead to information loss which results in a reduced MI between the two variables~\citep{tishby2000information, alemi2016deep, ji2018iic}.}
The equality holds when $g$ is \revision{an invertible} mapping between embedding space to $K$ categories, which is not the case for a linear projection $g$.

The conditional joint distribution of cluster labels
\revision{
\begin{equation}
p(Z,Z'\, | \, \img,\img') \, = \, g\big(\enc(\img)\big)\cdot \tr{g\big(\enc(T(\img))\big)}
\end{equation}
}
yields a $K \times K$ probability matrix for each $\img \in \Data_u$, $\img'=T(\img)$, and $T$ sampled from $\TransSet$. After marginalizing over the entire $\Data_u$ (or a large mini-batch in practice), the $K \times K$ joint probability distribution \revision{$\PP = p(Z,Z')$} can be estimated as
\begin{equation}
 \PP \ \approx \ \frac{1}{|\Data_u|\, |\TransSet|}\sum_{\img \in \Data_{u}} \sum_{T \in \TransSet} g\big(\enc(\img)\big)\cdot \tr{g\big(\enc(T(\img))\big)}.
 \label{equ:cluster_join}
\end{equation}
Using the definition of MI in (\ref{eq:mi-kl}), the proposed global MI loss can then computed from $\PP$ as follows:
\begin{equation}
 \lossMI^{\mr{global}}(\params; \Data_u) \ = \ -\MI(\PP) \ = \ -\sum_{c=1}^{K}\sum_{c'=1}^{K} \PP_{c,c'}\, \log\,\frac{\PP_{c,c'}}{\sum_{c=1}^{K}\PP_{c,c'}\cdot \sum_{c'=1}^{K}\PP_{c,c'}}
 \label{equ:mi_given_join}
\end{equation}
A high $\MI(\PP)$ means that the information of $T(\img)$ can be retrieved given $\img$ (i.e., low conditional entropy), which forces the encoder to learn transformation-invariant features and, more importantly, group together images with similar feature representations. 

\begin{figure}[!t]
 \centering
 \includegraphics[width=0.95\linewidth]{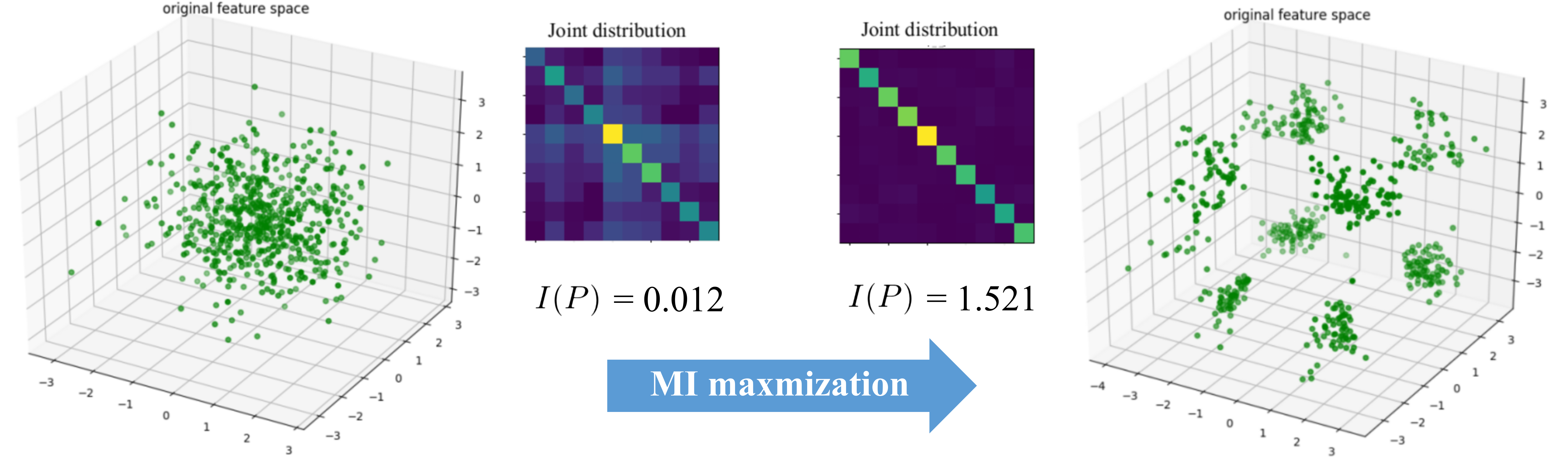}
 \caption{A toy example illustrating the effect of maximizing global MI. By increasing $\MI(\PP)$, randomly generated 3-D embedding points are effectively grouped into well-defined clusters.}
 \label{fig:mi_c_toy}
\end{figure}

To illustrate the clustering effect of $\lossMI^{\mr{global}}$, a simple example in 3-D feature space is presented in Fig.~\ref{fig:mi_c_toy}, where each randomly-generated point can be regarded as the three-dimensional embedding of an image. Optimizing $\lossMI^{\mr{global}}$ groups these embeddings into multiple clusters based on their relative positions. As a result, the joint distribution $\PP$ becomes confident with near-uniform values on the diagonal. This indicates that balanced clusters are formed and embedding points are pushed away from the decision hyperplanes defined by $g$. 

\subsection{Local mutual information loss} 

Our global MI loss focuses on the discriminative nature of encoder features, assuming that image-level contextual information can be captured. This may not be true for representations produced by decoder blocks. Given the features generated by the encoder, decoder blocks try to recover the  spatial resolution of features and produce densely-structured representations. Therefore, features from the decoder will also capture local patterns that determine the final segmentation output. Based on this idea, we propose a local MI loss $\lossMI^{\mr{local}}$ that preserves the local information of feature embeddings in the decoder.

\revision{Let $\psi^{(b)}(\img) = \dec^{(b)}\big(\enc(\img)\big) \in \Real^{C_b \times H_b \times W_b}$ be the feature map produced in the $b$-th decoder block for an unlabeled image $\img$. As described in Section \ref{sec:implementation}, each block is composed of a convolution and an upsampling operation}. This feature map has a reduced spatial resolution compared to $\img$ and segmentation output $\lbl$, and each of its feature vectors is a compact summary of a sub-region in the input image determined by the network's receptive field. Inspired by the fact that a region in an image shares information with adjacent, semantically-related ones, we maximize the MI between spatially-close elements of $\psi^{(b)}(\img)$. Denoting as $\big[\psi^{(b)}(\img)\big]_{i,j} \in \Real^C$  the feature vector located at position $(i,j)$ of the feature map, we define the neighbors of this vector using a set of displacement vectors $\Delta^{(b)} \subset \Int^2$:
\begin{equation}
\neigh^{(b)}_{i,j} \ = \ \big\{\big[\psi^{(b)}(\img)\big]_{i+p,j+q} \, | \, (p,q) \in \Delta^{(b)}\big\}.
\end{equation}
Furthermore, to make the decoder transformation invariant, we also enforce feature embeddings to have a high MI if they come from the same image under a data transformation $T \in \TransSet$. Note that unlike for the global MI loss, where the feature map is considered as a single representation vector, we now have to align the two embeddings in a same coordinate system. Hence, we need to compare $\big[\psi^{(b)}(T(\img))\big]_{i,j}$ with $\big[T(\psi^{(b)}(\img))\big]_{i+p,j+q}$, where $(p,q)$ is a displacement in $\Delta^{(b)}$. As before, we use a linear projection head $h$ to convert the feature map $\psi^{(b)}(\img)$ to a cluster assignment $h(\psi^{(b)}(\img)) \in [0,1]^{K_b \times H_b \times W_b}$. Since we want to preserve the spatial resolution of the feature map, $h$ is defined as a 1\,$\times$\,1 convolution followed by a softmax. Following \citep{peng2020mutual,ji2018iic}, we then compute a separate joint distribution $\PP^{(b)}_{p,q}$ for each displacement $(p,q) \in \Delta^{(b)}$:
\begin{equation}\label{eq:joint-local}
\PP^{(b)}_{p,q} \ \approx \ \frac{1}{|\Data_u|\,|\TransSet|\,|\Omega|}\sum_{\img \in \Data_u}\sum_{T\in \TransSet}\sum_{(i,j)\in \Omega}h\big([\psi^{(b)}(T(\img))]_{i,j}\big) \cdot \tr{h([T(\psi^{(b)}(\img))]_{i+p,j+q})} \\
\end{equation}
Note that the operation in (\ref{eq:joint-local}) can be computed efficiently with standard convolution operations. Finally, we obtain the local MI loss by averaging the MI over all decoder blocks $b \in \{1,\ldots,B\}$ and corresponding displacements:
\begin{equation}\label{eq:localMI_loss}
 \lossMI^{\mr{local}}(\params; \Data_u) \ = \ - \frac{1}{B} \sum_{b=1}^B \frac{1}{|\Delta^{(b)}|}
 \sum_{(p,q) \in \Delta^{(b)}}\!\! I\big(\PP^{(b)}_{p,q}\big) 
\end{equation}
where $\MI\big(\PP^{(b)}_{p,q}\big)$ is computed as in (\ref{equ:mi_given_join}). 


\subsection{Consistency-based loss}

As we will show in experiments, employing only the MI-based regularization losses may be insufficient to achieve optimal performance. This is in part due to the clustering nature of these losses: for two distributions conditionally independent given the same input image, MI is maximized if there is a deterministic mapping between clusters in each distribution such that they are equivalent. For example, permuting the cluster labels in one of the two distributions does not change their MI. 

To ensure the alignment of cluster labels throughout the network, we add a final loss term $\lossCons$ which imposes the network output at each pixel of an unlabeled image to remain the same under a set of transformations. In this work, we measure output consistency using the $L_2$ norm:
\begin{equation}
 \loss_{\mathrm{cons}}(\params; \Data_u) \ = \ \frac{1}{|\Data_u| \, |\TransSet| \, |\Omega|}\sum_{\img \in \Data_u}\sum_{T\in \TransSet}\sum_{(i,j)\in \Omega} \big\|f_{ij}(T(\img)) - T(f_{ij}(\img))\big\|_2^{2}
 \label{equ:consistency}
\end{equation}
This loss, which is typical to approaches based on transformation consistency, has been shown to boost segmentation performance in a semi-supervised setting \citep{bortsova2019semi}.

\section{Experimental setup}\label{sec:experiments}

\subsection{Dataset and metrics}

To assess the performance of the proposed semi-supervised method, we carried out extensive experiments on four clinically-relevant benchmark datasets for medical image segmentation: the Automated Cardiac Diagnosis Challenge (ACDC) dataset~\citep{bernard2018deep}, the Prostate MR Image Segmentation (PROMISE) 2012 Challenge dataset~\citep{litjens2014evaluation}, the Spleen sub-task dataset of the Medical Segmentation Decathlon Challenge~\citep{simpson2019large}, and the Multi-Modality Whole Heart Segmentation (MMWHS) dataset~\citep{zhuang2016multi}. These four datasets contain different image modalities (CT and MRI) and acquisition resolutions.

\paragraph{\textbf{ACDC dataset}} The publicly-available ACDC dataset consists of 200 short-axis cine-MRI scans from 100 patients, evenly distributed in 5 subgroups: normal, myocardial infarction, dilated cardiomyopathy, hypertrophic cardiomyopathy, and abnormal right ventricles. Scans correspond to end-diastolic (ED) and end-systolic (ES) phases, and were acquired on 1.5T and 3T systems with resolutions ranging from 0.70\,$\times$\,0.70 mm to 1.92\,$\times$\,1.92 mm in-plane and 5 mm to 10 mm through-plane. Segmentation masks delineate 4 regions of interest: left ventricle endocardium (LV), left ventricle myocardium (Myo), right ventricle endocardium (RV), and background. We consider the 3D-MRI scans as 2D images through-plane due to the high anisotropic acquisition resolution, and re-sample them to a fix space ranging of $1.0 \times 1.0\,\mathrm{mm}$. Pixel intensities are normalized based on the $1\%$ and $99\%$ percentile of the intensity histogram for each patient. Normalized slices are then cropped to 384\,$\times$\,384 pixels to slightly adjust the foreground delineation of the ground truth. For the main experiments, we used a random split of 8 fully-annotated and 167 unlabeled scans for training, and the remaining 25 scans for validation.
\revision {In another experiment, we also evaluate our model trained with a varying number of patient scans as labeled data.} A rich set of data augmentation was employed for both labeled and unlabeled images, including random crops of $224\,\times224\,$ pixels, random flip, random rotation within [$-$45, 45] degrees, and color jitter.

\paragraph{\textbf{Prostate dataset}} This second dataset focuses on prostate segmentation and is composed of multi-centric transversal T2-weighted MR images from 50 subjects. These images were acquired with multiple MRI vendors and different scanning protocols, and are thus representative of typical MR images acquired in a clinical setting. Image resolution ranges from 15\,$\times$\,256\,$\times$\,256 to 54\,$\times$\,512\,$\times$\,512 voxels with a spacing ranging from 2\,$\times$\,0.27\,$\times$\,0.27 to 4\,$\times$\,0.75\,$\times$\,0.75 mm$^3$. 2D images are sliced along short-axis and are resized to a resolution of $256\,\times\, 256$ pixels. A normalization is then applied on pixel intensity based on $1\%$ and $99\%$ percentile of the intensity histogram for each patient. We randomly selected 4 patients as labeled data, 36 as unlabeled data, and 10 for validation during the experiments. For data augmentation, we employ the same set of transformation as the ACDC dataset, except we limit the random rotation to [$-$10, 10] degrees.

\paragraph{\textbf{Spleen dataset}} The third dataset consists of patients undergoing chemotherapy treatment for liver metastases. A total of 41 portal venous phase CT scans were included in the dataset with acquisition and reconstruction parameters described in~\citep{simpson2019large}. The ground truth segmentation was generated by a semi-automatic segmentation software and then refined by an expert abdominal radiologist. Similar to the previous dataset, 2D slices are obtained by slicing the high-resolution CT volumes along the axial plane. Each slice is then resized to a resolution of 512\,$\times$\,512 pixels for the sake of normalization. To evaluate algorithms in a semi-supervised setting, we randomly split the dataset into labeled, unlabeled and validation image subsets, comprising CT scans of 6, 30, and 5 patients respectively. For data augmentation, we employ a random crop of $256\,\times256\,$ pixels, color jitter, random horizontal flip, and random rotation of [$-$10, 10] degrees.

\paragraph{\textbf{Multi-Modality Whole Heart Segmentation (MMWHS) dataset}} \revision{The last dataset includes 20 high-resolution CT volumes from 20 patients. The in-plane resolution is around $0.78 \times 0.78 \mathrm{mm}$ and the average slice thickness is 1.60 mm. Following the same protocol as for the ACDC dataset, we prepossessed and sliced three dimensional images into 2D slices with a fixed space ranging of $1.0 \times 1.0\,\mathrm{mm}$. All slices were then center-cropped to 256\,$\times$\,256 pixel. We randomly split the dataset into labeled (2 patients), unlabeled (13 patients) and validation (5 patients) sets, which were fixed throughout all experiments. We employ the same set of data augmentations as for ACDC. 
}

\revision{For all the datasets}, we used the commonly-adopted Dice similarity coefficient (DSC) metric to evaluate segmentation quality. DSC measures the overlap between the predicted labels ($S$) and the corresponding ground truth labels ($G$):
\begin{equation}
 \mathrm{DSC}(S,G) = \frac{2|S\cap G|}{|S|+|G|} 
\end{equation}
DSC values range between 0 and 1, a higher value corresponding to a better segmentation. In all experiments, we reconstruct the 3D segmentation for each patient by aggregating the predictions made for 2D slices and report the 3D DSC metric for the validation set.

\subsection{Implementation details}\label{sec:implementation}

\paragraph{Network and parameters} For all four datasets, we employ the same U-Net architecture comprised of 5 Convolution\,+\,Downsampling blocks in the encoder, 5 Convolution\,+\,Upsampling blocks in the decoder, and skip connections between convolutional blocks of same resolution in the encoder and decoder (see Fig. \ref{fig:diagram} for details). We adopted this architecture as it was shown to work well for different medical image segmentation tasks.

Network parameters are optimized using stochastic gradient descent (SGD) with the Adam optimizer. For all experiments, we applied a learning rate warm-up strategy to increase the initial learning rate of $1\times 10^{-7}$ for \revision{both ACDC and MMWHS}, $1\times 10^{-6}$ for Prostate and $1\times 10^{-6}$ for Spleen by a factor of 400 in the first 10 epochs and decreases it with a cosine scheduler for the following 90 epochs. We define an epoch as 300 iterations, each consisting of a batch of 4 labeled and 10 unlabeled images drawn with replacement from their respective dataset. The proposed MI-based regularization is applied to the feature embeddings generated in three different blocks: the last block of the decoder (\textbf{Conv5}) for the global MI loss, and the last two convolutional blocks from the decoder (\textbf{Upconv3} and \textbf{Upconv2}) for the local MI loss. In an ablation study, we measure the contribution of regularizing each of these embeddings on segmentation performance. 

We employ an array of five linear projectors, instead of a single projector, to project feature embeddings to a corresponding set of categorical distributions, and average the MI-based losses over these distributions. For the encoder, the projector head consists of a max-pooling layer to summarize context information, a linear layer and a softmax activation layer. On the other hand, for the decoder, we only use a 1$\times$1 convolution with softmax activation layer.
{As the proposed MI-based losses are computed on their output, the parameters of these projectors are also updated during training.}
We also tested projection head consisting of several layers with non-linearity, however this resulted in a similar performance but a higher variance. In the default setup of our method, we fixed the number of clusters to $K = 10$ for both the encoder and the decoder. In another ablation study, we show that a slightly greater performance can be achieved for a larger $K$, at the cost of increased computations. 

To balance the different regularization terms in (\ref{eq:totalLoss}), we used weights of $\lambda_1=0.1$, $\lambda_2=0.1$ and $\lambda_3=5$ for experiments on the ACDC \revision{and MMWHS} datasets, $\lambda_1=0.05$, $\lambda_2=0.05$ and $\lambda_3=10$ for the Prostate dataset, and $\lambda_1=0.05$, $\lambda_2=0.05$ and $\lambda_3=5$ for the Spleen dataset. These hyper-parameters were determined by grid search. We set the pool of transformations on unlabeled images ($\TransSet$) as random horizontal and vertical flips. For the local MI loss, we set the neighborhood size $\Delta$ to be 3$\times$3 for \textbf{Upconv3} and 7$\times$7 for \textbf{Upconv2}, corresponding to a regions of 3-5 mm in original image space depending on the resolution. We also tested our method with larger neighborhoods, however this increased computational cost without significantly improving accuracy.

\paragraph{\revision{Compared methods}} We compared our method against several baselines, \revision{ablation variants of our method} and recently-proposed approaches for semi-supervised segmentation:
\begin{itemize}[itemsep=1pt,topsep=2pt]
 \item \textbf{Full supervision}: We trained the network described above using the supervised loss $\lossSup$ on \textit{all} training images. This results in an upper bound on performance.
 \item \textbf{Partial Supervision}: A lower bound on performance is also obtained by optimizing $\lossSup$ only on \textit{labeled} images, ignoring the unlabeled ones. 
 
 \item \textbf{Mutual information}: \revision{This ablation variant of our method} consists in maximizing MI for intermediate feature embeddings while ignoring the consistency constraint on the output space (i.e., dropping $\lossCons$ in the loss).
 
 \item \textbf{Consistency regularization}~\citep{bortsova2019semi}: \revision{This second ablation variant}, which can be seen as the $\Pi$ model for image segmentation, imposes $\loss_{\mathrm{cons}}$ loss as the only regularization loss, without using $\lossMI^{\mr{global}}$ or $\lossMI^{\mr{local}}$. Only the output distribution space is regularized while embeddings from intermediate features are unconstrained.
 
 \item \textbf{Entropy minimization}~\citep{vu2019advent}: In addition to employing $\lossSup$ on labeled data, this well-known semi-supervised method minimizes the pixel-wise entropy loss of predictions made for unlabeled images. By doing so, it forces the network to become more confident about its predictions for unlabeled images. 
 To offer a fair comparison, we performed grid search on the hyper-parameter balancing the two loss terms, and report the score of the best found hyper-parameter.
 
 \item \textbf{Mean Teacher}~\citep{perone2019unsupervised}: This last approach adopts a teacher-student framework where two networks sharing the same architecture learn from each other. Given an unlabeled image, the student model seeks to minimize the prediction difference with the teacher network whose weights are a temporal exponential moving average (EMA) of the student's. We use the formulation similar to \citep{perone2019unsupervised} and quantify the distribution difference using $L_{2}$ loss. Following the standard practice, we fix the decay coefficient to be 0.999. The coefficient balancing the supervised and regularization losses is once again selected by grid search. 
\end{itemize}
All tested methods are implemented in a single framework, which can be found here: \paperurl.

\textbf{Additional experiments} To further assess the improvement on segmentation quality brought by the proposed global and local MI losses, we performed additional experiments on the ACDC dataset. We first compare our method against Mean Teacher for different amounts of labeled data. \revision{Second, we examine the sensitivity of our method to different regularization weights.} Third, we investigate the importance of features from different hierarchical levels of the network. Fourth, we evaluate the impact on segmentation quality of using a different number of clusters $K$ in projection heads of the proposed architecture. \revision{Finally, we show that the joint optimization of labeled and unlabeled data losses works better in our semi-supervised setting than a two-step strategy of pre-training the network on a clustering task using unlabeled data, and then fine-tuning it on the segmentation task with labeled data.}

\section{Experimental Results}\label{sec:results}

\begin{table}[!t]
\centering
\caption{Mean 3D DSC of tested methods on the ACDC, Prostate, Spleen and MMWHS datasets. RV, Myo and LV refer to the right ventricle, myocardium and right ventricle classes, respectively. Mutual information corresponds to our method without loss term $\lossCons$ and Consistency regularization corresponds to our proposed loss without $\lossMI^{\mr{global}}$ or $\lossMI^{\mr{local}}$. \revision{For ACDC, Prostate, Spleen and MMWHS, respectively $5\%$, $10\%$, $16.7\%$ and $13.3\%$ of training images are considered as annotated and the rest as unlabeled.} Reported values are averages (standard deviation in parentheses) for 3 runs with different random seeds.}\label{table:result}
\begin{footnotesize}
\setlength{\tabcolsep}{5pt}
 \begin{adjustbox}{max width=\textwidth}
\begin{tabular}{lcccccccc}
\toprule
& \multicolumn{4}{c}{\textbf{ACDC}} & & \\
\cmidrule(l{8pt}r{8pt}){2-5}
 & \bfseries RV & \bfseries Myo & \bfseries LV & \bfseries Mean & ~\textbf{Prostate}~ & \textbf{Spleen} & \textbf{MMWHS}\\
 \midrule
Full supervision & 87.64 (0.46) &	87.46 (0.15)	&93.55 (0.33)	&89.55 (0.29) & 87.70 (0.13) & 95.32 (0.70) & 88.91 (0.12)\\
\midrule
 Partial Supervision & 57.67 (1.54) & 69.68 (2.35) &	86.08 (1.15) & 71.14 (1.28) & 41.63 (2.41) & 88.20 (1.89) &48.50 (1.73)\\
Entropy min. & 56.69 (3.56)	&73.46 (1.53)	&86.80 (2.36)	&72.32 (1.22) & 55.47 (2.05) & 90.77 (0.92) &49.44 (1.43)\\
Mean Teacher & 80.04 (0.48)	& 81.81 (0.17)	& 90.44 (0.33)	& 84.10 (0.26) & 80.61 (1.63) & 93.12 (0.57) &55.57 (0.48) \\
\midrule
Ours (MI only) & 78.73 (0.82) & 79.38 (0.40) & 88.80 (0.66) & 82.30 (0.57) & 74.75 (1.89) & 92.46 (0.80) &50.66 (1.38) \\
Ours (Consistency only) & 75.21 (0.94) & 82.31 (0.19) & \textbf{91.91 (0.47)}	& 83.14 (0.44) & 77.92 (1.20) & 94.19 (0.62) &49.15 (0.77) \\
Ours (all) & \textbf{81.87 (0.54)} & \textbf{83.65 (0.26)}	& 91.76 (0.32)	& \textbf{85.76 (0.16)} & \textbf{81.76 (0.71)} & \textbf{94.61 (0.65)} &\textbf{55.75 (0.40)} \\
\bottomrule
\end{tabular}
\end{adjustbox}
\end{footnotesize}
\end{table}

\subsection{Comparison with the state-of-the-art}

Table \ref{table:result} reports the mean 3D DSC obtained by tested methods on the validation set of the ACDC, Prostate, Spleen and MMWHS datasets. While using limited labeled data in training (e.g., 5\% of the training set as labeled data for ACDC), large performance gaps are observed between partial and full supervision baselines, leaving space for improvements to the regularization techniques. Overall, all semi-supervised approaches tested in this experiment improved performance compared to the partial supervision baseline, showing the importance of also considering unlabeled data during training. Entropy minimization, the worse-performing semi-supervised baseline, yielded absolute DSC improvements of 1.20\%, 13.83\%, 2.57\% and 0.94\% for the ACDC, Prostate, Spleen and MMWHS datasets, respectively. Mean Teacher and Consistency regularization gave comparable results, both of them outperforming Entropy minimization by a large margin. This demonstrates the benefit of enforcing output consistency during learning, either directly or across different training iterations as in Mean Teacher. With respect to these strong baselines, the proposed method achieved a higher 3D DSC in all but one case (left ventricle segmentation in ACDC). When averaging performance over the RV, Myo and LV segmentation tasks of ACDC, our method obtains the highest mean DSC of 85.76\%, compared to 84.10\% for Mean Teacher and 83.14\% for Consistency regularization. These improvements are statistically significant in a one-sided paired t-test (p $<$ 0.01). \revision{The robustness of our method to the execution random seed (network parameter initialization, batch selection, etc.) can also be observed by the low standard deviation values obtained for all datasets and tasks.}

\begin{figure}
 \centering
 \setlength{\tabcolsep}{0.5pt}
 \renewcommand{\arraystretch}{.8}
 \begin{footnotesize}
 \begin{tabular}{cccccc}
 \acdc{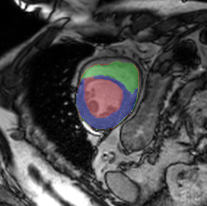} \\ 
 \acdc{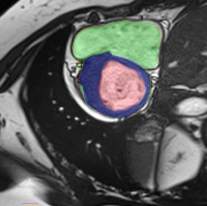} \\
 \acdc{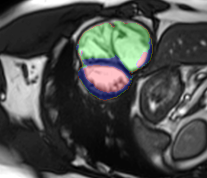} \\
 \prostate{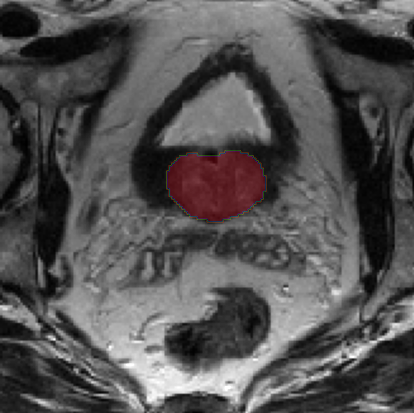} \\
 \prostate{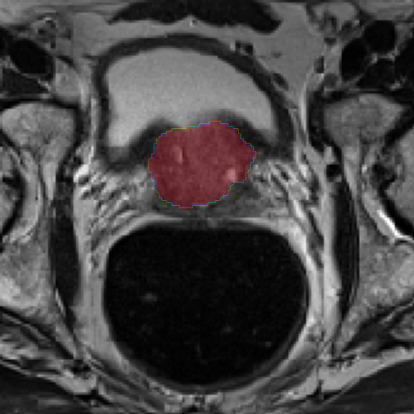} \\
 \spleen{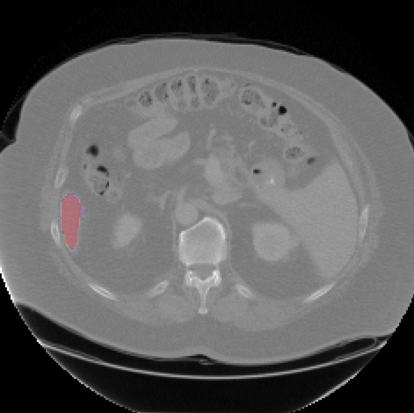} \\
 \spleen{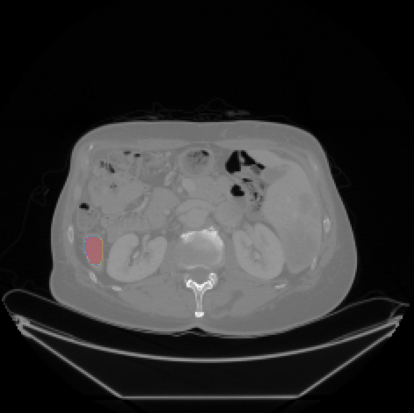} \\
 \mmwhs{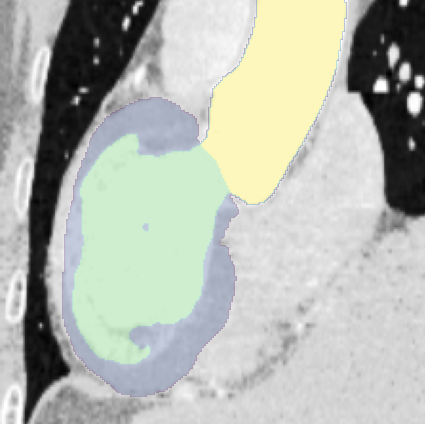} \\
 Ground truth & Partial Sup. & Mutual Info. & Consistency reg. & Mean Teacher & Our method
 \end{tabular}
 \end{footnotesize}
 \caption{Visual comparison of tested methods on validation images. \textbf{Rows 1--3}: ACDC; \textbf{Rows 4--5}: Prostate; \textbf{Rows 6--7}: Spleen; \textbf{Row 8}: MMWHS.}
 \label{fig:visual_inspect}
\end{figure}

The results in Table \ref{table:result} show that the combination of the MI-based and consistency-based losses in the proposed method are essential to its success. Considering only MI maximization (Mutual Information method) yields a mean DSC improvement of 11.16\% over the Partial Supervision baseline for ACDC, whereas performance is boosted by 12.00\% when also enforcing transformation consistency on the output. Similar results are obtained for the Prostate and Spleen datasets. As mentioned before, this could be explained by the fact that MI is invariant to label permutation, therefore a pixel-wise consistency loss such as $L_2$ is necessary to align these labels across different cluster projections of features. The performance of our method can be appreciated visually in Fig. \ref{fig:visual_inspect}, which shows examples of segmentation results for the tested methods. It can be seen that our method gives spatially-smoother segmentation contours that better fit those in the ground-truth. This results from regularizing network features both globally and locally. In contrast, only enforcing output consistency as in Mean Teacher and Consistency regularization leads to a noisier segmentation.

\begin{figure}[ht!]
 \centering
 \includegraphics[width=0.60\linewidth]{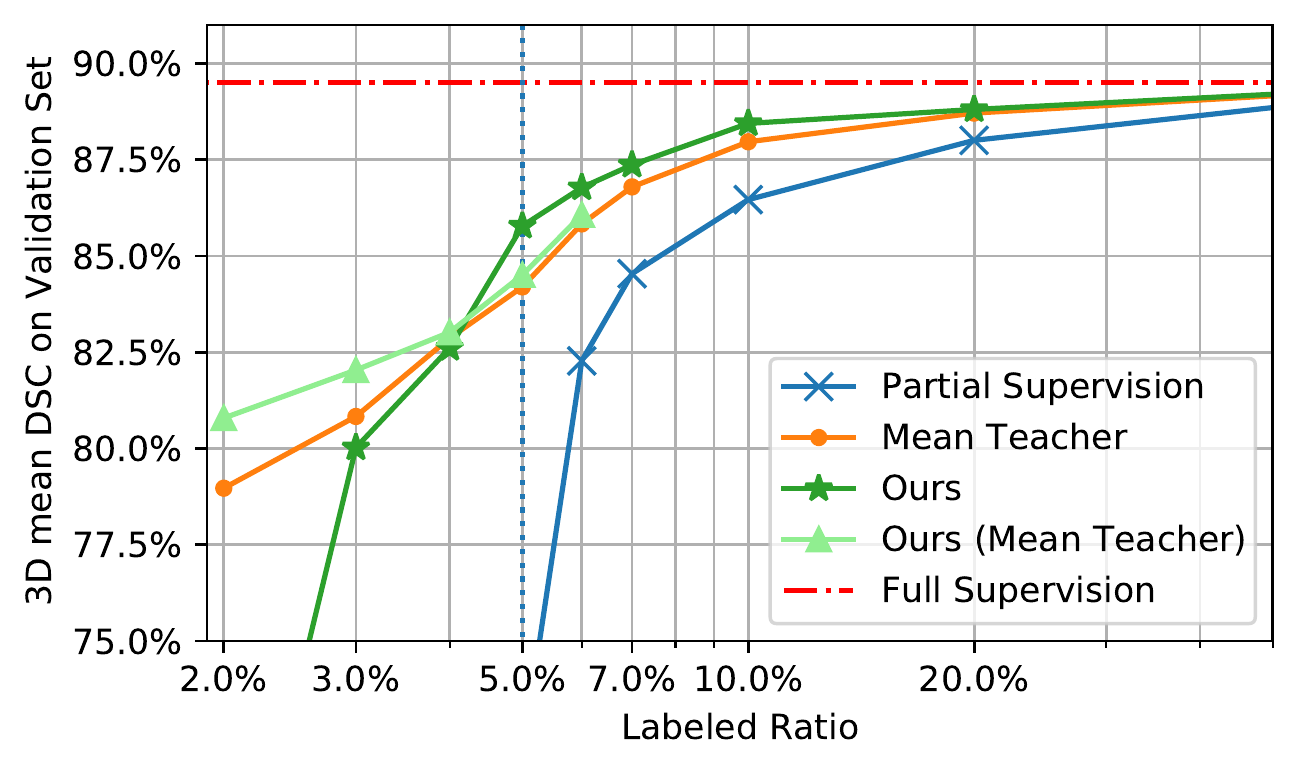}
 \caption{ACDC validation DSC versus various labeled data ratio for tested methods. It is clearly observed that our proposed method achieves higher performance compared with the state-of-the-art Mean Teacher in the regime of labeled ratio $\geq$ 5$\%$. For an extreme case where only $2-3\%$ data are provided with annotations, our enhanced adaption outperforms Mean Teacher.}
 \label{fig:visual_label_ratio}
\end{figure}

\subsection{Impact of labeled data ratio}

We further assess our method's ability to perform in a low labeled-data regime by training it with a varying number of labeled examples from the ACDC dataset, ranging from 2\% to 50\% of available training samples. As illustrated in Fig. \ref{fig:visual_label_ratio}, the proposed method (Dark green) offers a consistently better segmentation performance compared to Mean Teacher when over 5\% of training examples are annotated. By exploiting temporal ensembling, Mean Teacher (Orange) provides a more plausible segmentation when given an extremely limited amount of labeled data (less than $3\%$ of training samples). \revision{This can be attributed to the fact that, when trained with very limited labeled data, a single neural network is likely to overfit on those few examples and thus yield poor predictions for unlabeled images. Mean Teacher works well in this case as it exploits a separated Teacher network that distillates the knowledge of the student acquired at different training epochs, thereby implicitly smoothing the optimization and providing more a stable prediction on unlabeled images.}

Since the two methods are orthogonal, we can enhance our method by adapting it to the teacher-student framework of Mean Teacher. Toward this goal, we instead maximize the MI between feature embeddings of the teacher and the student, where the teacher's weights are computed as an expected moving average (EMA) of the student's. From Fig. \ref{fig:visual_label_ratio} (Light green), we see that this enhanced version of our method offers a good trade-off between Mean Teacher and our default model. While its performance is similar to Mean Teacher for a labeled data ratio of 4\% or more, it give a higher DSC when fewer annotated examples are provided. Thus, it improves the performance of Mean Teacher by 1.84\% when only $2\%$ of training samples are annotated.

\begin{table}[!t]
\centering
\caption{ACDC validation DSC of our method using feature embeddings from different network layers. \revision{5\% of training samples are considered as labeled.}}
\label{tab:impact_layer}
\begin{adjustbox}{max width=\textwidth}
\begin{tabular}{cccccccccc}
\toprule
\multicolumn{1}{c}{\bm{$\lossMI^{\mr{global}}$}} & 
\multicolumn{4}{c}{\bm{$\lossMI^{\mr{local}}$}} &
\multicolumn{5}{c}{\textbf{ACDC validation DSC}} \\
\cmidrule(l{8pt}r{8pt}){1-1}
\cmidrule(l{8pt}r{8pt}){2-5}
\cmidrule(l{8pt}r{8pt}){6-10}
\textbf{Conv5}& \textbf{Upconv5} & \textbf{Upconv4} & \textbf{Upconv3} & \textbf{Upconv2} & \textbf{RV}& \textbf{Myo} & \textbf{LV}& \textbf{Mean} & \textbf{Gain}\\ \midrule
\cmark & & & & & 76.29 & 81.83 & 90.65 & 82.92& 11.78 \\
& \cmark & & & & 76.34 & 82.61 & 91.68 & 83.54& 12.40 \\
& & \cmark & & & 80.04 & 81.39 & 90.00 & 83.81& 12.67 \\
 &
 &
 &
\cmark &
 &
\textbf{81.16} &
\textbf{83.30} &
\textbf{91.99} &
\textbf{85.48} &
\textbf{14.34} \\
& & & & \cmark & 77.65 & 81.96 & 90.90 & 83.50& 12.36 \\ \midrule
\cmark & \cmark & & & & 78.03 & 82.86 & 91.31 & 84.07& 12.93 \\
\cmark & & \cmark & & & 76.63 & 81.33 & 90.54 & 82.83& 11.69 \\
\cmark &
 &
 &
\cmark &
 &
\textbf{81.56} &
\textbf{83.14} &
\textbf{\underline{92.17}} &
\textbf{85.62} &
\textbf{14.48} \\
\cmark & & & & \cmark & 77.54 & 82.00 & 90.05 & 83.20& 12.06 \\ 
\midrule
\cmark & \cmark & \cmark & & & 79.88 & 82.72 & 91.51 & 84.70& 13.56 \\
\cmark & \cmark & & \cmark & & 78.24 & 82.94 & 91.67 & 84.28& 13.14 \\
\cmark & \cmark & & & \cmark & 77.58 & 82.27 & 90.15 & 83.33& 12.19 \\
\cmark & & \cmark & \cmark & & 79.80 & 82.21 & 90.90 & 84.30& 13.16 \\
\cmark & & \cmark & & \cmark & 79.11 & 83.17 & 91.18 & 84.49& 13.35 \\
\cmark &
 &
 &
\cmark &
\cmark &
\textbf{\underline{81.87}} &
\textbf{\underline{83.65}} &
\textbf{91.76} &
\textbf{\underline{85.76}} &
\textbf{\underline{14.62}} \\ 
\bottomrule
\end{tabular}
\end{adjustbox}
\end{table}
{
\color{black}

\subsection{Sensitivity to regularization loss weights}

We carried out experiments on the ACDC dataset to investigate the relative impact on performance of the loss terms in Eq. (\ref{eq:totalLoss}), as defined by weights $\lambda_{1}$, $\lambda_2$ and $\lambda_3$. To simplify the analysis, the weights controlling our global and local mutual information losses are set to the same value $\lambda_1=\lambda_2=\lambda_{\mathrm{MI}}$. The relative weight of the consistency loss, i.e. $\lambda_3$, is considered separately. We denote this weight as $\lambda_{\mathrm{con}}$ in the following results.

Table \ref{tab:impact_lambda} reports the mean 3D DSC performance on the ACDC dataset using $5\%$ of annotated data, for different combinations of $\lambda_{\mathrm{MI}}$ and $\lambda_{\mathrm{con}}$. We see that 
$\lambda_{\mathrm{MI}}$ has a significant impact on segmentation performance. In general, DSC increases when $\lambda_{\mathrm{MI}}$ goes from 0.01 to 0.1, and decreases rapidly when for larger values. In contrast, our method is less  sensitive to the choice of $\lambda_{\mathrm{con}}$, indicating that the proposed global and local MI-based losses contribute most to segmentation quality in a semi-supervised setting. 


\begin{table}[t!]
\centering
\caption{Mean DSC performance on the ACDC dataset given different $\lambda_{\mathrm{MI}}$ ($\lambda_{\mathrm{MI}}=\lambda_1=\lambda_2$) and $\lambda_{\mathrm{con}}$. \revision{5\% of training samples are considered as labeled.}}\label{tab:impact_lambda}
\begin{small}
\begin{tabular}{cccccc}
\toprule
 \multirow[b]{2}{*}{\textbf{$\lambda_{\mathrm{con}}$}} & 
 \multicolumn{5}{c}{\textbf{$\lambda_{\mathrm{MI}}=\lambda_1=\lambda_2$}} \\
\cmidrule(l{8pt}r{8pt}){2-6}
  & 0.01 &  0.05 & 0.1 & 0.5 & 1.0 \\
\midrule  
1.0 & 83.24\%&85.3\%&85.48\%&83.79\%&81.19\%\\
5.0 & 84.46\%&85.12\%&\textbf{85.76\%}&84.05\%&82.30\%\\
10.0 & 84.47\%&85.41\%&85.44\%&83.59\%&81.12\%\\
15.0 & 84.18\%&85.61\%&85.17\%&83.79\%&80.89\%\\
\bottomrule
\end{tabular}
\end{small}
\end{table}

}

\subsection{Impact of embedding layers}

The proposed MI-based losses regularize intermediate feature embeddings from both the encoder and decoder. The third experiment seeks to determine the impact of considering feature maps in different layers on results for the ACDC dataset. Since the global MI loss only uses features from the encoder's \textbf{Conv5} layer, we consider settings with and without this loss. On the other hand, the local MI loss regularizes features in four layers of the decoder: \textbf{Upconv2}-\textbf{Upconv5}. For our experiment, we test different combinations using a single or two of these layers in the local MI loss. Except for the selected features embeddings, the same training setting is used in all cases. \revision{Note that we set the neighborhood size $\Delta$ to be 1$\times$1 for both \textbf{Upconv5} and \textbf{Upconv4} as their resolution scales correspond to $1/8$ and $1/4$ of an input image.}

We observe from Table \ref{tab:impact_layer} that the choice of layers at which features are regularized has a noticeable impact results. Regularizing only encoder features (\textbf{Conv5}) in the global MI loss offers the smallest benefit. This may be due to the fact that segmentation requires learning the dense structure of an image, which is not well captured by the low-resolution features of the encoder. Conversely, highest improvements come from cases where \textbf{Upconv3} is selected in the local MI loss. The feature map in this layer has $1/2$ the resolution of the input image and, therefore, captures both global and local information. Overall, the best configuration is obtained with a combination of global regularization (\textbf{Conv5}) and local regularization (\textbf{Upconv3} and \textbf{Upconv2}), yielding a 14.62\% gain in DSC over the  Partial Supervision baseline.  

\subsection{Impact of cluster number $K$}
\label{sec: impact_k}

\begin{table}[t!]
\centering
\caption{Mean DSC performance on the ACDC dataset given different number of clusters $K$ for the encoder and decoder. \revision{5\% of training samples are considered as labeled.}}\label{tab:impact_k}
\begin{small}
\begin{tabular}{ccccc}
\toprule
 \multirow[b]{2}{*}{\textbf{Decoder $K$}} & \multicolumn{4}{c}{\textbf{Encoder $K$}} \\
\cmidrule(l{8pt}r{8pt}){2-5}
  & 2& 5 &  10 & 20 \\
\midrule  
2 & 84.37\%&84.66\%&84.58\%&84.56\%\\
5 & 84.70\%&85.43\%&85.86\%&86.05\%\\
10 & 84.86\%&85.33\%&85.76\%&85.91\%\\
20 &85.01\%&85.52\%&86.06\%&\textbf{86.32\%} \\
\bottomrule
\end{tabular}
\end{small}
\end{table}

A key component of our method is using auxiliary projectors to convert continuous feature representations to discrete cluster assignments. This encourages the grouping of semantically-related images/regions and enables the efficient computation of MI. As a result, the number of clusters $K$ at each layer may also impact performance: if $K$ is too small, image/region representations can only be grouped into a few discrete categories and, consequently, the network may fail to fully capture dependencies in the data. On the other hand, employing a very large $K$ requires having a large batch size and can result in high variance. 

In the next experiment, we tested different combinations of hyper-parameter $K \in \{2, 5, 10, 20\}$ for cluster assignments in the encoder (global MI loss) and decoder (local MI loss). Results of this experiments are summarized in Table \ref{tab:impact_k}. It can be seen that using a small $K=2$ for the encoder and decoder results in relatively low performance. Furthermore, increasing the number of clusters in either or both parts of the network generally improves segmentation quality. However, employing a larger $K$ also increases the computational cost of the method, especially for the local MI loss which relies on more expensive convolutional operations. On the whole, a value of $K=10$ offers a good trade-off between segmentation performance and run-time complexity. 

\subsection{Visualization of clusters}

\begin{figure}[!t]
 \centering
 \setlength{\tabcolsep}{0.5pt}
 \renewcommand{\arraystretch}{.8}
 \begin{small}
 \begin{tabular}{cccc}
\showfeaturespace{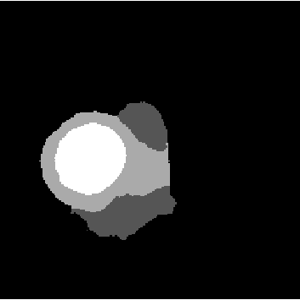} \\
\showfeaturespace{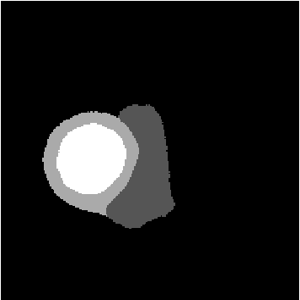} \\
\showfeaturespace{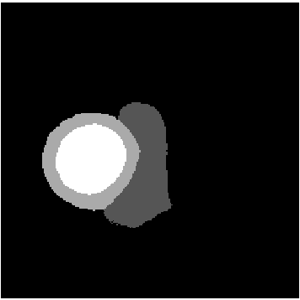} \\
\textbf{Input image} & \textbf{Upconv3} & \textbf{Upconv2} & \textbf{Prediction}
 \end{tabular}
 \end{small}
 \caption{Visual comparison of maximum activations taken from network decoder positions. \textbf{Top row}:  Partial Supervision. \textbf{Middle row}: Mean Teacher. \textbf{Bottom row}: Our method.}
 \label{fig:visual_feature_cluster}
\end{figure}

\begin{figure}[!t]
 \centering
 \setlength{\tabcolsep}{0.5pt}
 \renewcommand{\arraystretch}{.8}
 \begin{small}
 \begin{tabular}{ccc}
\includegraphics[width=0.33\linewidth]{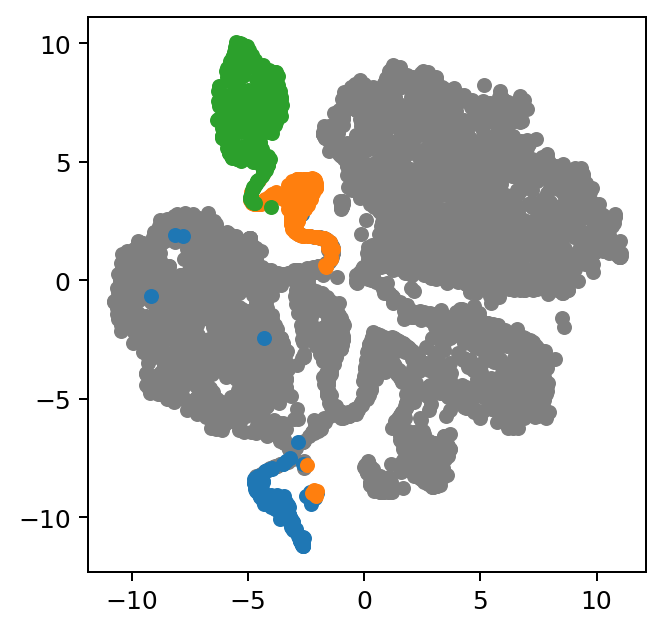} & 
\includegraphics[width=0.33\linewidth]{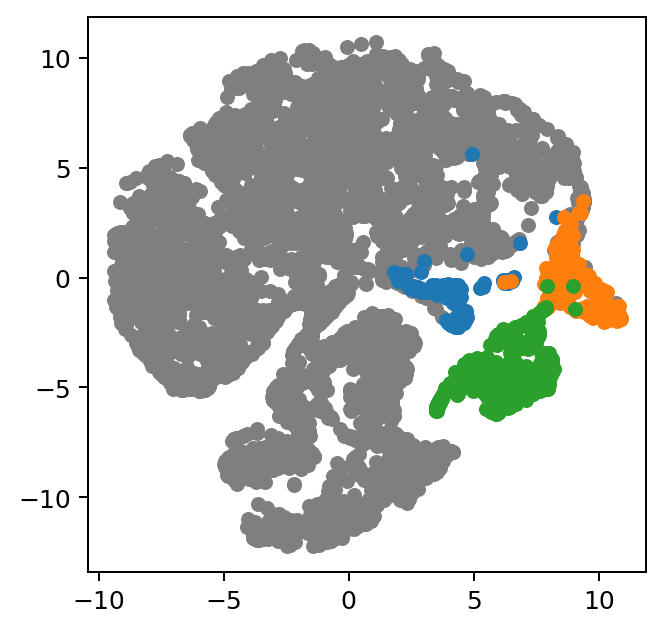} &
\includegraphics[width=0.33\linewidth]{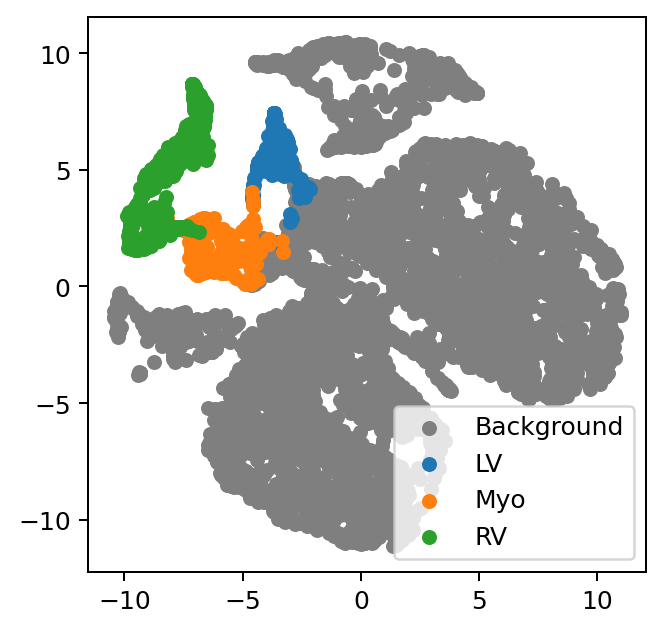} \\
 Partial Supervision & Mean Teacher & Ours 
 \end{tabular}
 \end{small}
 \caption{t-SNE plot on the ACDC validation set for different classes.}
 \label{fig:feature_tsne}
\end{figure}

Our method uses auxiliary projectors to map feature embeddings of corresponding images into categorical distributions. As mentioned before, this has a clustering effect where embeddings sharing similar semantic or structural information are grouped together while those with distinct information are pushed away. To illustrate this effect, we consider the ACDC dataset and plot in Fig. \ref{fig:visual_feature_cluster} the channel with highest activation at different positions of feature maps corresponding to decoder layers \textbf{Upconv3} and \textbf{Upconv2}. For visualization purposes, index values are mapped to the grey scale (min. index mapped to 0 and max. index to 255). The resulting channel map of our method is compared with those obtained using  Partial Supervision and Mean Teacher. Moreover, we give in Fig. \ref{fig:feature_tsne} the t-SNE plot of feature vectors at each position of the feature map in \textbf{Upconv2}, color-coded by the ACDC classes. 

We observe in Fig. \ref{fig:visual_feature_cluster} that  Partial Supervision outputs unrealistic predictions (rightmost column) and noisy feature activations (second and third columns). When trained with insufficient annotated data, a network can be misguided to learn noisy signals, such as local texture and geometric variability. In contrast, Mean Teacher and the proposed method produce segmentation maps similar to the ground truth. However, the feature activations of Mean Teacher appear noisier and less structured than those learned by our method. This confirms that regularizing only the output space results in a poor internal representation. In comparison, the feature activations of our method better correlate with the semantic information of ground truth labels. This result confirmed by the 2D t-SNE plot in Fig. \ref{fig:feature_tsne}, where nearby points corresponds to positions in the feature map with similar feature vectors. As can be seen, our method exhibits more compact clusters with less outliers compared to Partial Supervision and Mean Teacher. This spatial clustering effect leads to a smoother segmentation and reduces overfitting when training with limited supervision. 

\begin{figure}[!ht]
    \centering
    \includegraphics[width=0.6\linewidth]{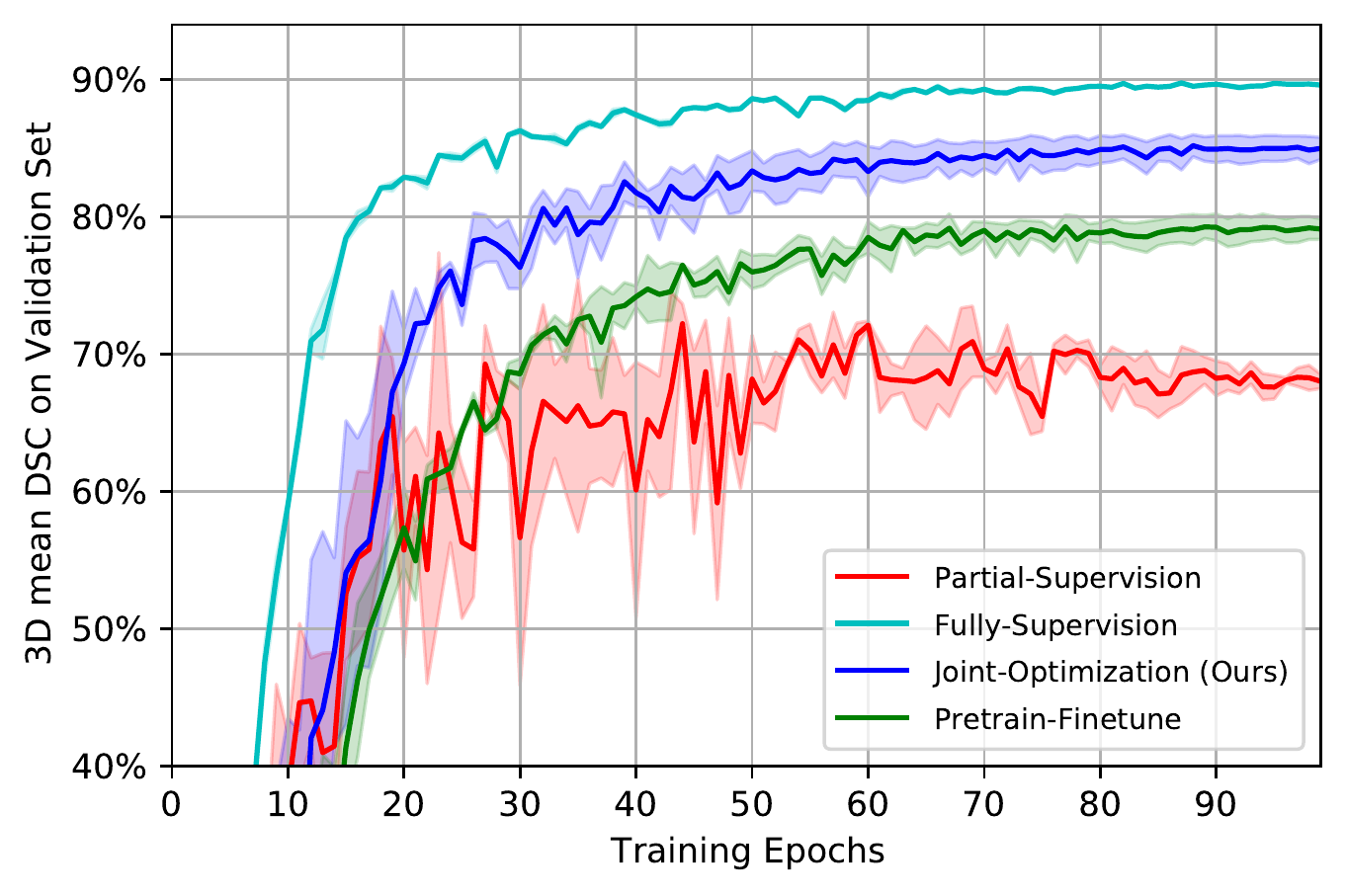}
    \caption{Validation mean DSC versus training epochs for the Pretrain-Finetune strategy and our Joint-Optimization method. Mean and standard deviation values are calculated from three independent runs.}
    \label{fig:pretrain-finetune}
\end{figure}

{\color{black}
\subsection{Joint optimization of supervised and unsupervised losses}
A significant difference between our method and \citep{ji2018iic} relates to how models are trained. The approach in \citep{ji2018iic} employs a two-stage training strategy where a feature representation is first obtained in an unsupervised way (clustering task), and then a mapping from clusters to segmentation labels is found based on a few labeled examples. In contrast, our method jointly optimizes both the supervised loss and unsupervised regularization terms during the whole training process. 

The next experiment on the ACDC dataset is designed to validate the advantage of joint optimization compared to the two-stage training procedure, called Pretrain-Finetune in the following results. For this experiment, we use a similar setup as in previous experiments (e.g., $5\%$ of training set as annotated examples) but make the following changes. For the first stage, we optimize a randomly-initialized segmentation network on \textit{all} images using Eq. (\ref{eq:totalLoss}), without the supervised loss term of Eq. (\ref{eq:cross-entropy}). By doing so, the network tries to learn a meaningful feature representation without annotations. In the second stage, we apply the supervised loss of Eq. (\ref{eq:cross-entropy}) only on labeled images, enabling the network to fine-tune its representation and propagate acquired knowledge to the segmentation output space. For a fair comparison, we once again performed a grid search to select $\lambda_1$, $\lambda_2$ and $\lambda_3$ for this unsupervised setting, and report performance obtained with the best set of hyper-parameters.

The evolution of performance, in terms of average DSC on the ACDC validation set, is shown in Fig. \ref{fig:pretrain-finetune}. The mean and standard deviation are reported from three independent runs with different random seeds. We see that the Pretrain-Finetune strategy helps stabilize the training process and boosts segmentation performance by nearly $5.10\%$ over Partial-Supervision. This confirms the ability of the proposed loss to learn useful representations when trained in an unsupervised setting. Nevertheless, our Joint-Optimization method outperforms this two-stage strategy by a significant margin, achieving the best validation score. This improvement is due to the fact that involving the labeled data into MI-based optimization helps the network find a more robust representation, thus improving the segmentation performance in an scarce-annotation setting. 
}

\section{Discussion and conclusion}

We presented a novel semi-supervised method for medical images segmentation which regularizes a network by maximizing the MI between semantically-related feature embeddings, both globally and locally. The proposed global MI loss encourages the encoder to learn a transformation-invariant representation for unlabeled images. On the other hand, the local MI loss captures high-order dependencies between spatially-related embeddings, and preserves structure under perturbations of the input. By combining these two MI-based losses with a consistency term that promotes the alignment of cluster labels across different feature embeddings, the network can be effectively trained with limited supervision. We applied the proposed method to four challenging medical segmentation tasks with few annotated images. Experimental results showed our method to outperform recently-proposed semi-supervised approaches such as Mean Teacher and Entropy minimization, offering segmentation performance near to full supervision. 


Standard loss functions for segmentation consider the prediction for different pixels as independent. An important advantage of our MI regularization losses is taking into consideration the structured nature of segmentation. Towards this goal, we maximize the MI on intermediate feature embeddings by using auxiliary projectors that map these continuous representations to a categorical distribution. While this provides an efficient way to estimate MI and promotes the grouping of semantically-related representations, other approximation techniques could also be explored. A possible alternative is adversarial contrastive learning \citep{bose2018adversarial}, which employs an adversarially-learned sampler to find a reduced set of hard negative samples. Reducing the number of negative samples required to estimate MI could make approaches based on contrastive learning better-suited for the segmentation of large images and 3D scans. Another way to enhance the proposed method would be to incorporate priors on the distribution of segmentation labels. By maximizing MI, the proposed method indirectly favors balanced sizes for the segmented regions. When segmenting regions of very different sizes, better results could be achieved by constraining the marginal distribution of outputs \citep{hu2018imsat}. As future work, we could also validate the proposed method on multi-modal images, and large-scale segmentation benchmarks such as Cityscapes~\citep{cordts2016cityscapes}. 



\acks{
We acknowledge the support of the Natural Sciences and Engineering Research Council of Canada (NSERC), and thank NVIDIA corporation for supporting this work through their GPU grant program.
}

\ethics{The work follows appropriate ethical standards in conducting research and writing the manuscript, following all applicable laws and regulations regarding treatment of animals or human subjects.}

\coi{We declare we don't have conflicts of interest.}





\vskip 0.2in
\bibliography{biblio}

\end{document}